\documentclass[10pt,twocolumn,letterpaper]{article}

\usepackage{wacv}
\usepackage{times}
\usepackage{epsfig}
\usepackage{graphicx}
\usepackage{amsmath}
\usepackage{amssymb}

\usepackage{amsfonts}
\usepackage{algorithm}
\usepackage{algpseudocode}

\def\wacvPaperID{234} 

\wacvfinalcopy 

\ifwacvfinal
\def\assignedStartPage{1} 
\fi

\ifwacvfinal
\usepackage[breaklinks=true,bookmarks=false]{hyperref}
\else
\usepackage[pagebackref=true,breaklinks=true,colorlinks,bookmarks=false]{hyperref}
\fi

\ifwacvfinal
\else
\fi

\begin{document}

\title{Network Generalization Prediction \\for Safety Critical Tasks in Novel Operating Domains}


\author{
Molly O'Brien\textsuperscript{1}\\
{\tt\small molly@jhu.edu}
\and
Mike Medoff \textsuperscript{2}\\
{\tt\small mmedoff@exida.com}
\and 
Julia Bukowski\textsuperscript{3}\\
{\tt\small julia.bukowski@villanova.edu}
\and 
Greg Hager\textsuperscript{1}\\
{\tt\small hager@cs.jhu.edu}
\and
\textsuperscript{1} Department of Computer Science, Johns Hopkins University, Baltimore MD 21218\\
\textsuperscript{2}exida LLC,  Sellersville PA 18960\\
\textsuperscript{3}Department of Electrical and Computer Engineering,  Villanova University, Villanova, PA 19085 \\
}

\maketitle

\begin{abstract}
It is well known that Neural Network (network) performance often degrades when a network is used in novel operating domains that differ from its training and testing domains. This is a major limitation, as networks are being integrated into safety critical, cyber-physical systems that must work in unconstrained environments, e.g., perception for autonomous vehicles.  Training networks that generalize to novel operating domains and that extract robust features is an active area of research, but previous work fails to predict what the network performance will be in novel operating domains. We propose the task \textbf{Network Generalization Prediction}: predicting the expected network performance in novel operating domains. We describe the network performance in terms of an interpretable Context Subspace, and we propose a methodology for selecting the features of the Context Subspace that provide the most information about the network performance. We identify the Context Subspace for a pretrained Faster RCNN network performing pedestrian detection on the Berkeley Deep Drive (BDD) Dataset, and demonstrate Network Generalization Prediction accuracy within $5\%$ or less of observed performance. We also demonstrate that the Context Subspace from the BDD Dataset is informative for completely unseen datasets, JAAD and Cityscapes, where predictions have a bias of $10\%$ or less. 

\end{abstract}


\section{Introduction}
Deep Neural Networks (networks) are being integrated into commercial, safety critical, autonomous systems that operate in unconstrained environments, e.g., perception for autonomous vehicles. When a network is deployed in an unconstrained environment, the operating domain, i.e., the distribution of context features that describe the network's environment, can change significantly from the testing domain, i.e., the distribution of context features that describe the test data. Safety critical systems are regulated by international functional safety standards, e.g., ISO 26262 for the automotive industry, IEC 61508 for electronics and software. Functional safety standards leverage various techniques to verify the safety of software, including requirement specification, i.e., linking required system behavior to specific code modules, white box testing, i.e., testing specific inputs that cover all branches or behavior in the code, and code inspection and review to identify human error. These techniques are  challenging or impossible to apply directly to networks, e.g., labeled data is used to implicitly specify the correct behavior in supervised learning, networks are black box systems, and network weights cannot be manually inspected to identify failure cases. 

New techniques are needed to bridge the gap between the high performance of deep networks and the verification required for safety critical systems. 
In particular, the ability to predict how a network's performance will change in a novel operating domain can enable verifying the required level of performance before a network is deployed, we denote this task Network Generalization Prediction. We propose a methodology for Network Generalization Prediction for networks trained via supervised learning. 
Our contributions are as follows: 

\begin{enumerate}
\item We introduce the concept of a Context Subspace, a low-dimensional space, encoding the context features most informative about the network performance. 
\item We propose a greedy feature selection algorithm for identifying the Context Subspace by 1) ranking the context features by the information they provide about the network loss, and 2) selecting the subspace dimensionality that leads to accurate Network Generalization Prediction. 
\item We leverage a Context Subspace for accurate Network Generalization Prediction for pedestrian detection in diverse operating domains, with a prediction error from $0.5\%$ to $2\%$ for not safety critical pedestrians (pedestrians not in the road), and a prediction error from $2\%$ to $5\%$ for safety critical pedestrians (pedestrians in the road). 
\item We demonstrate that the Context Subspace identified for the Berkeley Deep Drive Dataset can be used to predict pedestrian recall in completely unseen datasets, the JAAD and Cityscapes Datasets, with a prediction bias of $10\%$ or less. 
\end{enumerate}


\section{Background}
\label{sec:citations}

\subsection{Network Dependability}
Avizienis et al. defined software dependability as ``the ability to deliver service that can justifiably be trusted," where dependability encompasses availability, reliability, safety, integrity, and maintainability\cite{avizienis2004basic}. 
To describe the dependability of a learned model, O'Brien et al. defined ML Dependability as ``the probability that a model will succeed when operated under specified conditions"\cite{obrien2019dependable}. 
Cheng et al. proposed that Robustness, Interpretability, Completeness, and Correctness contribute to a network's Dependability \cite{cheng2018towards}.  Ponn et al. trained a random forest to predict whether a network would detect a pedestrian, based on pedestrian attributes; they denote this task Detection Performance Modeling\cite{ponn2020identification}. Where Detection Performance Modeling predicts whether one specific object will be detected, Network Generalization Prediction predicts the expected network performance for a given operating domain, described by a distribution of context features.

\subsection{Network Generalization}
It has been shown that underspecification causes network performance to degrade when deployed in operating conditions different from the training and testing conditions\cite{d2020underspecification}. The WILDS benchmark was released to provide datasets with ``in-the-wild" distribution shifts between the training and test data \cite{koh2020wilds}. Subbaswamy et al. propose to evaluate a model's robustness to distribution shifts with one fixed evaluation set \cite{subbaswamy2020evaluating}. 
 Common techniques to improve network generalization include extracting features robust to changing conditions\cite{taghanaki2020jigsaw}, \cite{kim2019learning}, zero or few-shot learning \cite{xian2018zero}, \cite{wang2019survey}, and identifying when an input is outside the network's training distribution \cite{liang2017enhancing}, \cite{Hsu_2020_CVPR}.




\subsection{Feature Selection}\label{rw:fs}
Feature selection algorithms aim to select a subset of the available features, typically to use the features as input to train a model for a given task. Feature selection algorithms can be classified as filter methods, i.e., features are scored according to their association with the task label, wrapper methods, i.e., features are selected to minimize task error, and embedded methods, i.e., features are selected in the model training process \cite{cai2018feature}. 
The Mutual Information \cite{kraskov2004estimating} is often used in filter methods to measure the information between a given feature and the desired label \cite{vergara2014review}. 
As exhaustive feature selection search is typically intractable, greedy feature selection algorithms are often used \cite{khanna2017scalable}, \cite{tsamardinos2019greedy}, \cite{jiao2021greedy}.  Note, greedy feature selection is related to matching pursuit in the sparse approximation literature \cite{tropp2007signal} and has applications in compressed sensing \cite{braun2015info}.

\begin{figure}[t]
\begin{center}
\includegraphics[width=0.8\linewidth]{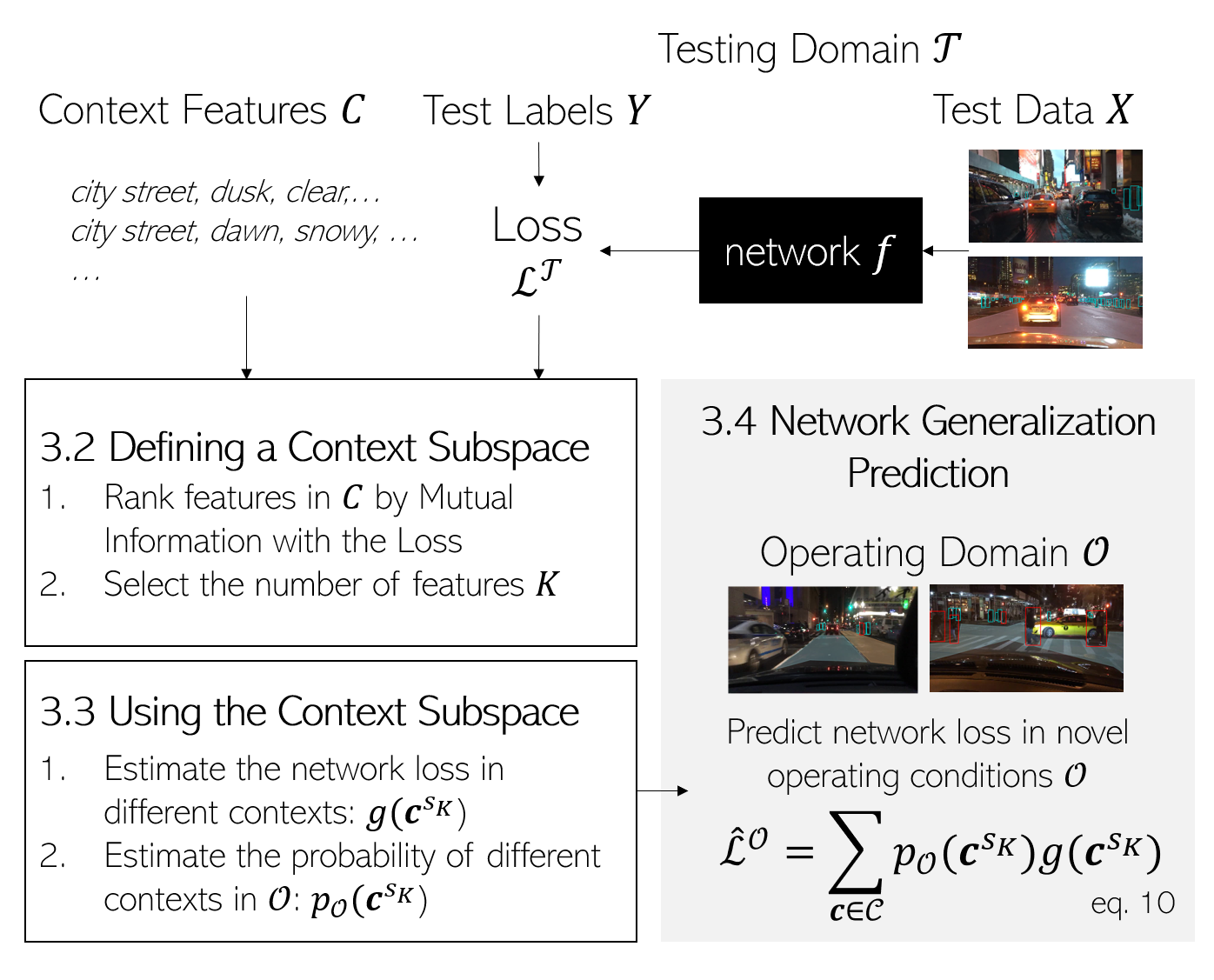}
\end{center}
   \caption{Overview of Network Generalization Prediction.}
\label{fig:methods_overview}
\end{figure}

\section{Methods}
\label{sec:methods}

\subsection{Problem Formulation}
It is well known that in supervised learning, a network, $f$, is trained to produce a label, $y_i$, from data, $x_i$, and a loss function, $l(f(x_i), y_i)$, is used to drive training. In Network Generalization Prediction, we are \textbf{not} training $f$. Instead, we aim to predict the performance of a fixed network $f$, trained via supervised learning, when deployed in an operating domain, $\mathcal{O}$, that differs from the testing domain, $\mathcal{T}$, see Figure \ref{fig:methods_overview}.  The performance of $f$ is measured using test data, $X = \{x_i\}_{i=1}^ N$, and test labels, $Y = \{y_i\}_{i=1}^ N$, via a loss function $L = \{l(f(x_i), y_i)\}_{i=1}^N$, where the elements of $L$ are assumed to be discrete and bounded, e.g., an object detection flag, whether a safety criteria was satisfied, or a discretized  classification error. 

$\mathcal{T}$ is described via $J$ context features, $\textbf{C} = \{ \textbf{c}_i \}_{i=1}^N$, where $\textbf{c}_i$ indicates a $J$ dimensional context vector associated with $x_i$. Context features, e.g., image brightness, weather, or robot speed, can be categorical or numerical; numerical features are assumed to be discrete or discretized. It is possible for multiple test samples to map to the same context, i.e., $\textbf{c}_i = \textbf{c}_j, i \neq j$.  $p_{\mathcal{T}}(\textbf{c})$ denotes the probability of encountering $\textbf{c}$ in $\mathcal{T}$.
$\mathcal{O}$ is described by the probability of encountering $\textbf{c}$ in $\mathcal{O}$, $p_{\mathcal{O}}(\textbf{c})$. In many practical applications, the likelihood of encountering a context may be known without annotated data, e.g., there is a $25\%$ chance of snow in Boston, etc. Note, labeled test data from $\mathcal{O}$ is not required. We assume that while the distribution of contexts shifts between the testing and operating domains, i.e., $p_{\mathcal{T}}(\textbf{c}) \neq p_{\mathcal{O}}(\textbf{c})$, the expected network performance in context $\textbf{c}$ is stable for both the testing and operating domains.
Table \ref{tab:notation} describes the Notation used in the Methods Section.


As is typical, we approximate the posterior expected loss in $\mathcal{T}$, $\mathcal{L}^\mathcal{T}$, using the empirical loss: 


\begin{equation}
\mathcal{L}^\mathcal{T} = E[l(f(X), Y)]  = \frac{1}{N} \sum_{i=1}^N l(f(x_i), y_i)
\end{equation}
We define $g(\textbf{c}) = E[l(f(X), Y | \textbf{c})]$.
 Let $\mathbb{I}(\textbf{a},\textbf{b})$ be an indicator function that is equal to $1$ if $\textbf{a} = \textbf{b}$ and $0$ otherwise. $g(\textbf{c})$ can be computed as:
\begin{equation}\label{eq:gck}
g(\textbf{c}) =  \frac{\sum_{i=1}^N {\mathbb{I}(\textbf{c}_i, \textbf{c}) * l(f(x_i), y_i)} }{\sum_{i=1}^N {\mathbb{I}(\textbf{c}_i, \textbf{c})}}
\end{equation}
 $\mathcal{L}^\mathcal{T}$ can equivalently be computed as:
\begin{equation}
\mathcal{L}^\mathcal{T} =  \sum_{ \textbf{c} \in \textbf{C}} p_{\mathcal{T}}(\textbf{c}) g(\textbf{c}) 
\end{equation}
Likewise, we can now express the Network Generalization Prediction, $\hat{\mathcal{L}}^{\mathcal{O}}$, as:
\begin{equation}\label{eq:ngp}
\hat{\mathcal{L}}^{\mathcal{O}} =  \sum_{\textbf{c} \in \textbf{C}} p_{\mathcal{O}}(\textbf{c}) g(\textbf{c})
\end{equation}
This formulation holds theoretically for any number of context features $J$. However, as $J$ grows linearly, computing Eqn. \ref{eq:ngp} requires exponentially more test samples to cover every possible $\textbf{c} \in \textbf{C}$. Thus, we introduce the Context Subspace, $\textbf{C}^{S_K}$, a low-dimensional space, encoding the context features most informative about the network performance.


\begin{table}
\begin{center}
\begin{tabular}{c l}
\\ & \textbf{Notation}   \\
$X = \{x_i\}_{i=1}^N$ & The Test Data \\
$Y = \{y_i\}_{i=1}^N$ & The Test Labels \\
$f$ & The trained network \\
$L = \{l(f(x_i), y_i)\}_{i=1}^N$ & The Test Set Loss \\ 

\hline 
$\textbf{C}$   & The context features \\
$\textbf{c} \in \textbf{C}$ & A context vector \\
$g(\textbf{c})$ & The expected loss of $f$ in  $\textbf{c}$\\
$\textbf{C}^{S_K}  $  & The Context Subspace\\
\hline  
$\mathcal{T}$ & The Testing Domain \\
$\mathcal{O}$ & The Operating Domain \\
$p_{\mathcal{T}}(\textbf{c})$, $p_{\mathcal{O}}(\textbf{c})$ & The probability of $\textbf{c}$ in $\mathcal{T}$, $\mathcal{O}$ \\
$\mathcal{L}^\mathcal{T}$  &The observed loss in  $\mathcal{T}$\\ 
$\hat{\mathcal{L}}^\mathcal{O}$ &The predicted loss in  $\mathcal{O}$
\end{tabular}
\end{center}
\caption{Notation.}\label{tab:notation}
\end{table}

\subsection{Defining a Context Subspace}\label{sec:define_context}
We are interested in selecting the $K$ context features that provide the most information about the network loss, to include these features in  $\textbf{C}^{S_K}$.
Let $S_K = \{s_k \}_{k=1}^K$ be the indices of context features of interest and $\textbf{C}^{S_K} = \{C^{s_k}\}_{k=1}^K$, where $ C^{s_k} = \{c_i^{s_k}\}_{i=1}^N$ are the annotated attributes for each example in the test set for context feature ${s_k}$.  
To select the context features to include in $\textbf{C}^{S_K}$, we 1) rank the context features by how much information they provide about the network loss, 2) select the $\textbf{C}^{S_K}$ dimensionality $K$ to enable accurate Network Generalization Prediction.  

\subsubsection{Ranking Context Features}\label{sec:rank}
Recall, the Mutual Information is often used to rank features in filter feature selection algorithms and is computed as $I(L,C^j)$ for loss $L$ and context feature $C^j$:
\begin{equation}
I(L,C^j) = \sum_{\ell \in L} \sum_{c \in C^j} p(\ell, c) log(\frac{p (\ell, c)} {p(\ell) p(c)} )
\end{equation}
where $p(\ell, c)$ indicates the joint probability of $\ell$ and $c$, and $p(\ell)$ and $p(c)$ indicate the marginal probabilities for $\ell$ and $c$, respectively. 
The Interaction Information is a generalization of the Mutual Information to $K$ features.  
The Interaction Information between $L$ and the context features $C^{s_1}, ..., C^{s_K}$ is defined as:
\begin{multline}
I(L, C^{s_1}, ..., C^{s_K}) 
= I(L, C^{s_1}, ..., C^{s_{K-1}}) \\ - I(L, C^{s_1}, ..., C^{s_{K-1}}|C^{s_K}) 
\end{multline}
For two features, this becomes:
\begin{equation}
I(L, C^{s_1}, C^{s_2}) = I(L, C^{s_2}) - I(L, C^{s_2}|C^{s_1}) 
\end{equation}
Where $I(L, C^{s_2}|C^{s_1}) $ can be computed as:
\begin{multline}
I(L, C^{s_2}|C^{s_1})  = \sum_{\ell \in L} \sum_{c_2 \in C_2} \sum_{c_1 \in C_1}  p(\ell, c_2, c_1) \\ 
\times  log \left( \frac{p(\ell, c_2, c_1)}{p(\ell, c_1) p(c_2, c_1)} \right)
\end{multline}
The computational complexity of $I(L, C^{s_1}, ..., C^{s_K})$ grows combinatorially with $K$. We are interested in ranking the context features by the Interaction Information, but computing the exact Interaction Information becomes intractable as $K$ grows.
To make computation tractable, we propose $\Delta I(L, C^{s_1}, ..., C^{s_K})$ to approximate how much more information including context feature $C^{s_K}$ in the Context Subspace provides about $L$. 
\begin{equation}
\Delta I(L, C^{s_1}, ..., C^{s_K})= I(L, C^{s_K}) - \sum_{k=1}^{K-1} I(C^{s_k}, C^{s_K})
\end{equation}
Intuitively, $\Delta I(L, C^{s_1}, ..., C^{s_K})$ subtracts the redundant information in $C^{s_K}$, $ \sum_{k=1}^{K-1} I(C^{s_k}, C^{s_K}) $, from the information it provides about the loss, $I(L, C^{s_K})$. Note that the computational complexity of computing $\Delta I(L, C^{s_1}, ..., C^{s_K})$ grows linearly with $K$. Like the Interaction Information, $\Delta I(L, C^{s_1}, ..., C^{s_K})$ can be positive or negative. In Appendix \ref{app:derivation}, we show that for independent features in the Context Subspace, $\Delta I(L, C^{s_1}, C^{s_2})$ approaches $I(L, C^{s_1}, C^{s_2})$ as $C^{s_2}$ approaches perfect information on $L$. We propose a greedy algorithm to iteratively select the $K$ most informative features from the context, see Algorithm \ref{greedy_alg}.
\begin{algorithm}[t]
\caption{Greedy $\Delta I$ Context Selection}
\label{greedy_alg}
\begin{algorithmic}[1]
\State $S_K = \{\}$
\For{ $k = 1:K $}
\State $s_k^* \leftarrow argmax_j [I(C^j, L) - \sum_{s_k \in S_K} I(C^j, C^{s_k})] $ \\ $ \quad       \quad \forall j \in J \setminus S_K$
\State $S_K = S_K \cup s_k^*$
\EndFor
\State $g(\textbf{c}^{S_K}) = E[l(f(X), Y | \textbf{c}^{S_K})]$
\end{algorithmic}
\end{algorithm}

\subsubsection{Selecting the Context Subspace Dimensionality}\label{sec:select_k}

Selecting the number of features, $K$, to include in $\textbf{C}^{S_K}$ is not trivial: including more features can lead to a more descriptive $\textbf{C}^{S_K}$ but can also lead to many untested contexts in $\textbf{C}^{S_K}$. To select $K$, we compute the expected prediction error for a given subspace dimensionality, $\epsilon_K$. Using the $K$ most informative context features, $g(\textbf{c}^{S_K}) = E[l(f(X), Y | \textbf{c}^{S_K})]$ can be computed according to Eqn. \ref{eq:gck}.
where $\textbf{c}^{S_K}$ is a $K$ dimensional feature vector in $\textbf{C}^{S_K}$. 
We iteratively compute the prediction error within the test set, $\epsilon_K$, to estimate the expected prediction error $\tilde{\epsilon}_K$, see Algorithm \ref{select_k_alg}. First, we randomly partition the test set into a $fit$ set and a $val$ set: $X^{fit}$, $Y^{fit}$, $\textbf{C}^{fit}$ with $N^{fit}$ samples and $X^{val}$, $Y^{val}$, $\textbf{C}^{val}$ with $N^{val}$ samples respectively.  We estimate $g^{fit}(\textbf{c}^{S_K})$ using the $fit$ set. We compute the observed loss from the $val$ set, $\mathcal{L}^{val}$. Let $p^{val}(\textbf{c}^{S_K})$ indicate the probability of encountering context $\textbf{c}^{S_K}$ in $\textbf{C}^{val}$. The prediction error, $\epsilon_K$, is the difference between the observed validation loss, $\mathcal{L}^{val}$, and the predicted validation loss using $g^{fit}(\textbf{c}^{S_K})$. This procedure can be iterated multiple times, and the subsequent $\epsilon_K$'s averaged, to estimate the expected prediction error, $\tilde{\epsilon}_K $, for different random $fit$ and $val$ partitions of the test set. We select the $K$ that minimizes $\tilde{\epsilon}_K $. 

\begin{algorithm}[h]
\caption{Context Subspace Dimensionality Selection}
\label{select_k_alg}
\begin{algorithmic}[1]
\State $\tilde{\epsilon}_K = \{\}$
\For{ $K = 1:J $}
\State $\epsilon_Ks = []$
\For{$iteration$}
\State split test set into $fit$ and $val$ set
\State $g^{fit}(\textbf{c}^{S_K}) = E[l(f(X^{fit}), Y^{fit} | \textbf{c}^{S_K})]$
\State $\mathcal{L}^{val} =  \frac{1}{N^{val}} \sum_{i=1}^{N^{val}} l(f(x^{val}_i), y^{val}_i)$
\State $\epsilon_K =  | \mathcal{L}^{val} -  \sum_{\textbf{c}^{S_K} \in \textbf{C}^{S_K}} p^{val}(\textbf{c}^{S_K}) g_k^{fit}(\textbf{c}^{S_K}) |$
\State $\epsilon_Ks.append(\epsilon_K)$
\EndFor
\State $\tilde{\epsilon}_K = mean(\epsilon_Ks)$
\EndFor
\State $K \leftarrow argmin_K \quad \tilde{\epsilon}_K$
\end{algorithmic}
\end{algorithm}

$\tilde{\epsilon}_K $ measures the expected prediction error within $\mathcal{T}$. When the context is informative about the loss, we expect $\tilde{\epsilon}_K $ to decrease as $K$ increases until an optimal $K^*$ is reached, then $\tilde{\epsilon}_K $ will begin to rise as $K$ increases and there are many untested contexts. If $\tilde{\epsilon}_K $ is flat or increasing as $K$ increases, it indicates that the context features available are not informative about the loss.

After we have ranked the context features and selected the number of features to include in the subspace, we can form $\textbf{C}^{S_K}$. The $K$ most informative context features form the axes of the subspace. Recall, we assumed the context features are categorical or numerical and discrete, this yields a finite set of context partitions, $\textbf{c}^{S_K} \in \textbf{C}^{S_K}$. 

\subsection{Using the Context Subspace}\label{sec:use_csk}
We use $\textbf{C}^{S_K}$ to describe the expected network loss in different contexts, $g(\textbf{c}^{S_K})$, and to describe the probability of encountering a context in the operating domain, $p_\mathcal{O}(\textbf{c}^{S_K})$. We can compute $g(\textbf{c}^{S_K})$ using Eqn. \ref{eq:gck}, note we use the entire test set to compute $g(\textbf{c}^{S_K})$ once we have selected the subspace dimensionality $K$. Recall, we do not assume to have labeled test data in $\mathcal{O}$, but we do assume to know $p_\mathcal{O}(\textbf{c}^{S_K})$.  Individual context feature probabilities can be multiplied to obtain a joint probability distribution if the context feature probabilities are assumed to be independent. 


\subsection{Network Generalization Prediction}\label{sec:network_gen_pred}
We can now perform Network Generalization Prediction, where $\hat{\mathcal{L}}^{\mathcal{O}}$ is the predicted loss in $\mathcal{O}$:
\begin{equation}\label{eq:ngp_csk}
\hat{\mathcal{L}}^{\mathcal{O}} =  \sum_{ \textbf{c}^{S_K} \in \textbf{C}^{S_K}} p_{\mathcal{O}}(\textbf{c}^{S_K}) g(\textbf{c}^{S_K})
\end{equation}
Recall, we selected a small number of informative context features so that it would be practical to describe the unique contexts $\textbf{c}^{S_K} \in \textbf{C}^{S_K}$, but there may be untested contexts in $\textbf{C}^{S_K}$. For conservative predictions, we assume the maximum loss in untested contexts. The maximum loss may correspond to a binary failure or a large expected error.  Leveraging $\mathbf{C}^{S_K}$ renders Network Generalization Prediction practical for interestingly complex applications, like perception for autonomous vehicles.


\section{Experimental Results}
\subsection{Pedestrian Detection Generalization}
Perception for autonomous vehicles is an active area of research, and systems that use deep networks to detect and avoid obstacles, like pedestrians, while driving are commercially available. Some of these commercial systems can be used in any driving conditions, at the user's discretion, and the operating domains can vary significantly in terms of the lighting conditions, e.g., daytime compared to night, road conditions, e.g., clear weather compared to rainy or snowy weather, and obstacle density, e.g., a residential street compared to a restricted access highway. It would be impractical for autonomous vehicle developers to test a perception system in every possible operating domain, but it is also imperative to know whether it is safe to use a perception system in a given operating domain. We perform experiments analogous to an autonomous vehicle developer: we test a fixed network in one testing domain, $\mathcal{T}$, and predict the network's performance in novel operating domains, where the distribution of context features vary significantly from $\mathcal{T}$. Our goal is to accurately predict the observed network performance when the network is used in a novel operating domain, $\mathcal{O}$.

We test a pretrained Faster RCNN \cite{ren2015faster} object detector for pedestrian detection, where the objects detected as $person$ are used as pedestrian detections. In our analysis, we consider pedestrians whose ground truth bounding box area is $\ge 300$ pixels. We evaluate the network performance at the pedestrian level. Pedestrians correctly detected with an $IoU > 0.5$ and a confidence score $> 0.5$ are assigned a loss of $0$; pedestrians that are not detected are assigned a loss of $1$\footnote{We are predicting the network's recall. We do not assign a loss for false positive detections; this same methodology can be used to predict network precision if that is of interest. We focus on recall because failing to predict a pedestrian who is truly present in the scene is a higher safety risk than trying to avoid a pedestrian who is not present. }. Pedestrians in images with multiple people are considered independently; images with no pedestrians present are not assigned any loss. 

\begin{figure*}[t]
\begin{center}
\includegraphics[width=0.8\linewidth]{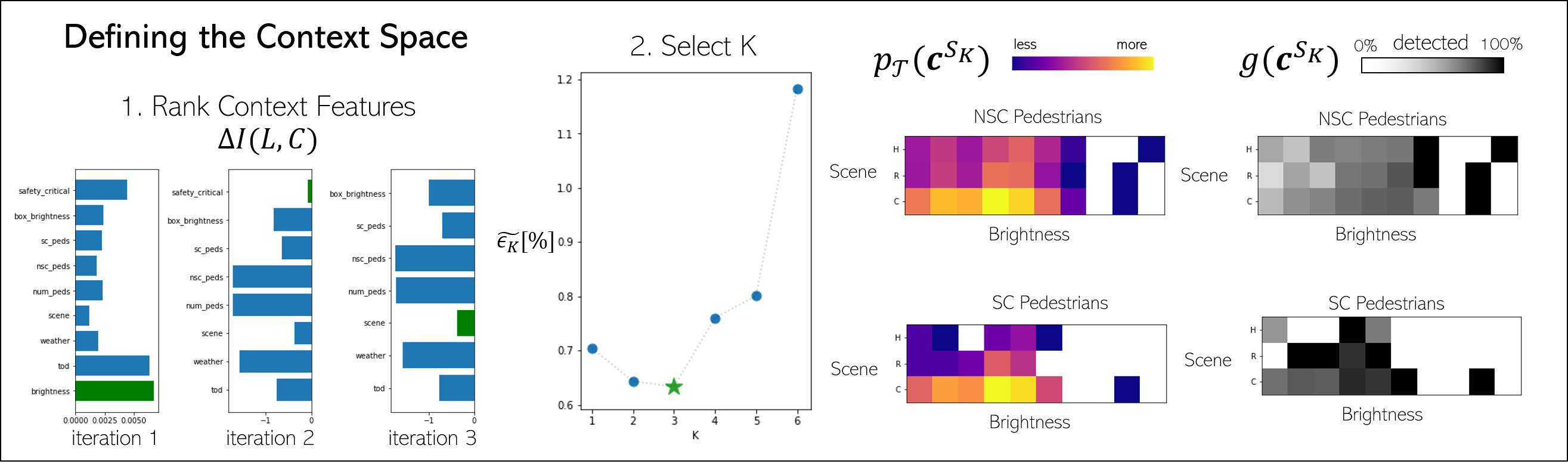}
\end{center}
   \caption{Defining the Context Subspace. 1) Rank Context Features: The $\Delta I(L,C)$ between different context features and the loss in the BDD Test Set for the first three rounds of Algorithm \ref{greedy_alg}. Note that in iteration one, $\Delta I(L,C)=I(L, C)$ so the features' scores are non-negative. 2) Select K: We estimate the expected prediction error for different Context Subspace dimensionalities, $K$, and choose the dimensionality with the lowest expected prediction error: in this case, $K=3$. We form the Context Subspace with the three most informative context features: brightness, safety critical flag, and the scene type. Right: heatmaps of the probability of encountering a context in the testing domain, $p_\mathcal{T}(\textbf{c}^{S_K})$, and the expected network loss in different contexts, $g(\textbf{c}^{S_K})$. X-axis: brightness (dark to bright from left to right). Y-axis: (top to bottom) scene type `H' highway, `R' residential, `C' city street. Separate heatmaps shown for NSC and SC pedestrians.}
\label{fig:ped_context}
\end{figure*}

The Berkeley Deep Drive (BDD) Dataset  \cite{yu2018bdd100k} was recorded across the continental US and includes data from varying times of day (daytime, dawn/dusk, or night), weather conditions (clear, partly cloudy, overcast, rainy, foggy, or snowy), and scene types (city street, residential, or highway). BDD images are of size $720 \times 1280$. We use $10,000$ images from the BDD Dataset for testing, denoted the BDD Test Set. We use the remaining $70,000$ images in the BDD Dataset, denoted the BDD Operating Set, to define novel operating domains.  The BDD Test Set and BDD Operating Set correspond to the BDD ``Validation" and ``Train" folds, respectively.

\subsection{Defining the Context Subspace}
We evaluate the network performance at the pedestrian level; therefore, context features are assigned to individual pedestrians. We do not know a priori which pedestrian attributes are informative about the network loss, so we include all available context features. The BDD dataset includes metadata on the image time of day, weather, and scene type.  We include the metadata as context features. We also include the image brightness and the pedestrian bounding box brightness. We define the road(s) to be the safety critical (SC) region(s) in the images. Pedestrians in the road are labeled SC, pedestrians outside the road, e.g., on the sidewalk, are labeled not safety critical (NSC).   The road is defined using the drivable area annotations. Whether a pedestrian is SC, denoted the safety critical flag, is included as a context feature. To capture information about the obstacle density in the scene, we include the total number of pedestrians, the number of SC pedestrians, and the number of NSC pedestrians  in the image as context features.

 \subsubsection{Ranking Context Features}
 We use Algorithm \ref{greedy_alg} to rank the context features by how much information they provide about the network loss. When computing the mutual information for a numerical feature with more than 10 unique values, we uniformly partition the feature into 10 discrete bins. Categorical features are labeled discretely with their assigned labels. See Figure \ref{fig:ped_context} for the $\Delta I$ computed for the first three iterations of Algorithm \ref{greedy_alg}. The 6 most informative features were found to be: 1) image brightness, 2) safety critical flag, 3) scene , 4) number SC pedestrians, 5) time of day, and 6) bounding box brightness.

\subsubsection{Selecting the Context Subspace Dimensionality}
To select the number of features to include in the Context Subspace, we compute $\tilde{\epsilon}_K$ for values of $K$ from $1$ to $6$. For each dimensionality, $K$, we compute $\epsilon_K$ 50 times  
by randomly partitioning the test data into $50\%$ for fitting $g(\textbf{c}^{S_K})$ and $50\%$  for validation. We select the $K$ with the minimum expected prediction error $\tilde{\epsilon}_K$ over the $50$ iterations.  $K=3$ was found to be optimal, with an average prediction error of $0.63\%$, see Figure \ref{fig:ped_context} center. We subsequently define the Context Subspace with three dimensions: 
1) image brightness, 2) safety critical flag, and 3) scene.

The image brightness is a continuous feature; we uniformly partition the image brightness into 10 bins. The safety critical flag and the scene type are discrete and categorical features with 2 and 3 possible values, respectively. This results in a Context Subspace, $\textbf{C}^{S_K}$, with $60$ discrete contexts, $\textbf{c}^{S_K}$.

\begin{figure*}[t]
\begin{center}
\includegraphics[width=0.8\linewidth]{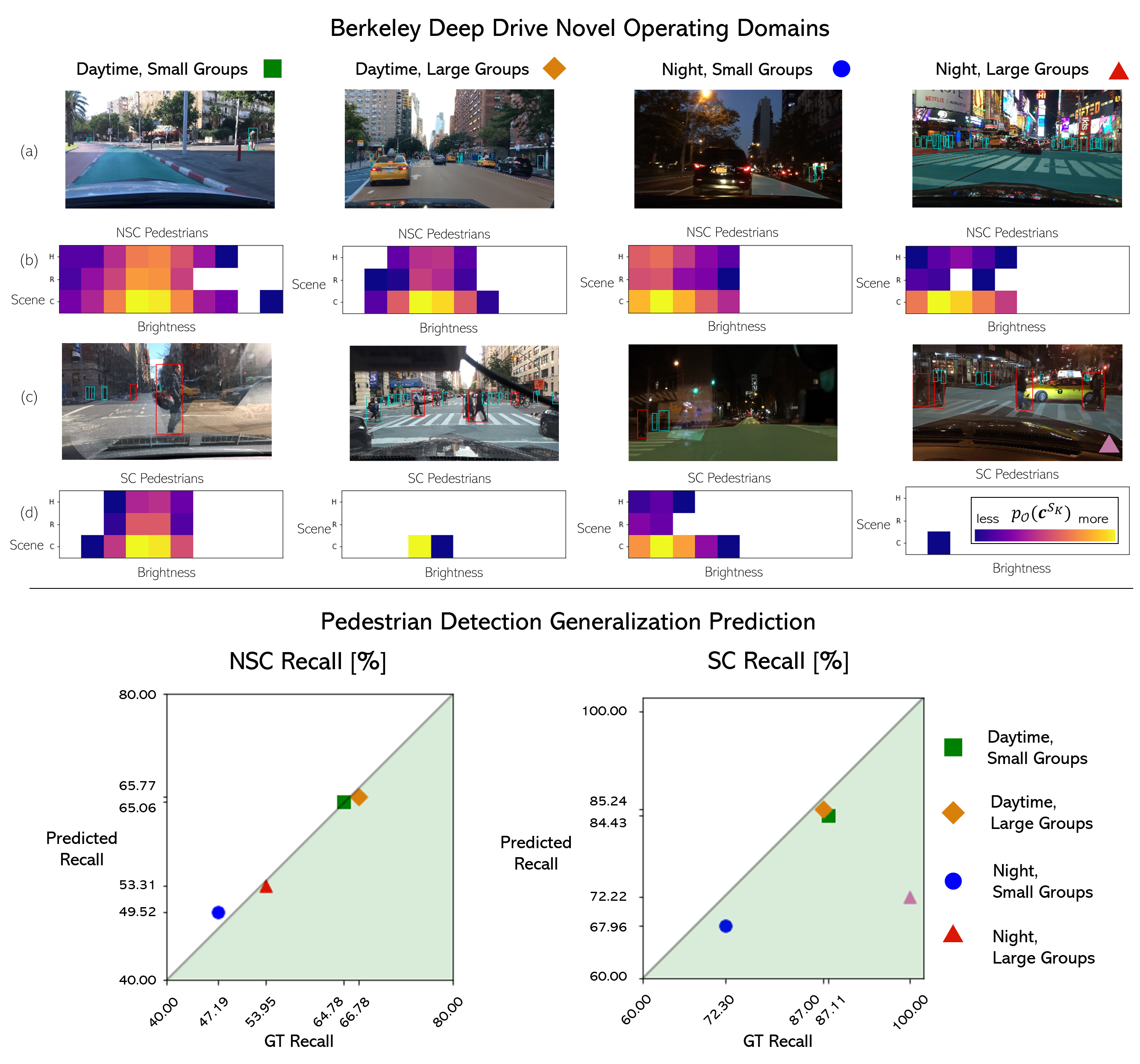}
\end{center}
   \caption{Top: BDD Novel Operating Domains. (a) Sample images of NSC pedestrians in the operating domain, NSC pedestrians outlined in cyan. Drivable area shown in transparent color. (b) NSC pedestrian $p_\mathcal{O}(\textbf{c}^{S_K})$. (c) Sample images of SC pedestrians in the operating domain, SC pedestrians outlined in red. Drivable area shown in transparent color. (d) SC pedestrian $p_\mathcal{O}(\textbf{c}^{S_K})$. Bottom: Pedestrian Generalization Prediction Results. NSC pedestrian recall and SC pedestrian recall are shown separately. X-Axis: Ground Truth (GT) recall in the operating domain $\mathcal{O}$. Y-Axis: predicted recall. Perfect predictions would fall on the diagonal line. Predictions in the shaded region are conservative, i.e., the predicted recall is less than the GT recall. }
\label{fig:bdd_preds}
\end{figure*}

\subsection{Using the Context Subspace}\label{res:use_csk}
We use $\textbf{C}^{S_K}$ to estimate the expected network loss in a context, $g(\textbf{c}^{S_K})$, and to describe the probability of encountering a context in $\mathcal{O}$, $p_{\mathcal{O}}(\textbf{c}^{S_K})$, see Figure \ref{fig:ped_context} right.  For all tested contexts, $g(\textbf{c}^{S_K})$ is computed according to Eqn. \ref{eq:gck}. All untested contexts are assigned an expected loss of $1$, i.e., a $100\%$ chance of failing to detect a pedestrian. The BDD Operating Set is used to define four novel operating domains: 1) daytime, small groups; 2) daytime, large groups; 3) night, small groups; and 4) night, large groups. The time of day annotated in the images was used to assign ``daytime'' or ``night''.  The SC and NSC pedestrians are considered independently. Pedestrians in images with fewer than $5$ (N)SC pedestrians are categorized as small groups; pedestrians in images with $5$ or more (N)SC pedestrians are categorized as large groups, i.e., in an image with 2 SC pedestrians and 15 NSC pedestrians, the SC pedestrians would be labeled `small group' and the NSC pedestrians would be labeled `large group'. We compute $p_{\mathcal{O}}(\textbf{c}^{S_K})$ for each $\mathcal{O}$ by counting the number of pedestrians that fall into each $\textbf{c}^{S_K} \in \textbf{C}^{S_K}$ and dividing by the total number of pedestrians. 

\subsection{Pedestrian Detection Generalization Prediction}
We predict the network loss in the novel operating domains defined in \ref{res:use_csk} using Eqn. \ref{eq:ngp_csk}. The heatmaps of $p_\mathcal{O}(\textbf{c}^{S_K})$ in Figure \ref{fig:bdd_preds} illustrate that the novel operating domains are significantly different from each other and the testing domain, see $p_\mathcal{T}(\textbf{c}^{S_K})$ in Figure \ref{fig:ped_context}. Our network loss is equivalent to the fraction of pedestrians that are not detected by the network; we convert the predictions into the predicted network recall by subtracting the fraction of pedestrians that are not detected from 1, see Figure \ref{fig:bdd_preds}. We then pass the BDD Operating Set through the network; the observed recall is computed as the fraction of pedestrians that were correctly detected. Figure \ref{fig:bdd_preds} illustrates that our predictions are accurate with Network Generalization Prediction accuracy between $0.5\%$ and $2.5\%$ for NSC pedestrian recall and $2\%$ and $5\%$ for SC pedestrian recall. All the SC predictions underpredict the observed recall; this demonstrates that our predictions are conservative.  Note, the only prediction with significant error is for night, large group SC pedestrians. Only one image in the BDD Operating Set falls into this category, so the observed performance is based on minimal data.

\begin{figure*}[t]
\begin{center}
\includegraphics[width=0.8\linewidth]{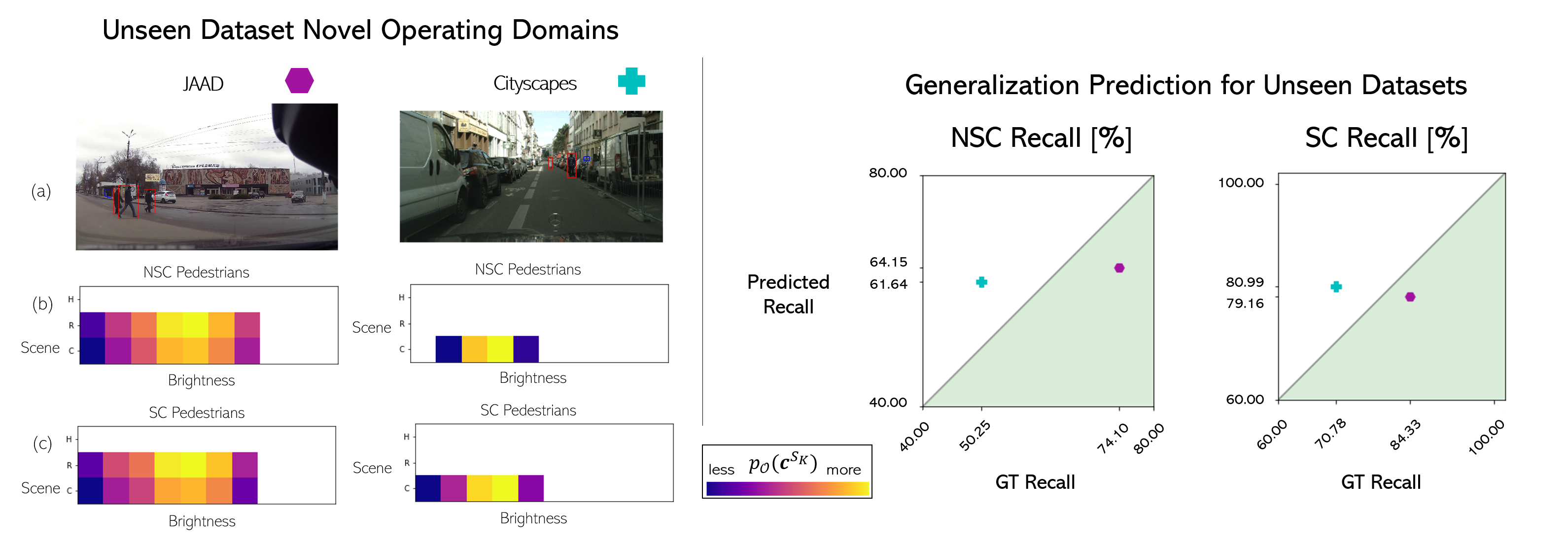}
\end{center}
   \caption{Left: Unseen Dataset Novel Operating Domains. (a) Sample images from the unseen datasets. NSC pedestrians outlined in blue. SC pedestrians outlined in red. (b) NSC pedestrian $p_\mathcal{O}(\textbf{c}^{S_K})$. (c) SC pedestrian $p_\mathcal{O}(\textbf{c}^{S_K})$. Right: Generalization Prediction for Unseen Datasets. NSC pedestrian recall and SC pedestrian recall are shown separately. X-Axis: Ground Truth (GT) recall in the dataset. Y-Axis: predicted recall. Perfect predictions would fall on the diagonal line. Predictions in the shaded region are conservative, i.e., the predicted recall is less than the GT recall. }
\label{fig:novel_datasets}
\end{figure*}

\subsection{Generalization Prediction for Unseen Datasets}
As a preliminary study, we investigate whether the Context Subspace, $\textbf{C}^{S_K}$, defined using the BDD Test Set and the network loss, $g(\textbf{c}^{S_K})$, estimated from the BDD Test Set provide information about completely unseen datasets. Unseen datasets include shifts in the context feature distributions, as well as changes in camera parameters and physical setup that are not captured by the test set. As such, we expect predictions for unseen dataset to contain bias, i.e., the prediction error for an unseen dataset will have a consistent non-zero offset. We are interested in determining the magnitude of this prediction bias to evaluate the usefulness of Network Generalization Prediction across datasets. We perform Network Generalization Prediction for the JAAD Dataset \cite{rasouli2017ICCVW}, and the Cityscapes Dataset with the gtFine labels \cite{Cordts2016Cityscapes}, see Figure \ref{fig:novel_datasets} for sample images. For both datasets, the (N)SC pedestrian image brightness distribution is computed from the images. 

The JAAD Dataset was recorded in North America and Europe; it includes primarily daytime images from residential and city streets in varying weather conditions. JAAD images are of size $1080 \times 1920$. For the JAAD Dataset, we sampled images every three seconds from the videos to limit temporal correspondence between frames; this resulted in 1,031 images.  Pedestrians in the road were manually annotated as SC, all others were labeled NSC. Scene annotations are not available for the JAAD dataset. To estimate the probability distribution of scenes, the scene type was annotated for a subset of 100 images, we assume the distribution holds for the entire dataset. The marginal (N)SC image brightness distributions and scene type distribution are multiplied to obtain the joint probability distributions for the JAAD Dataset.

The Cityscapes Dataset contains $3,475$ images recorded in 50 cities across Germany in the daytime during fair weather conditions.  Cityscapes images are of size $1024 \times 2048$. We defined the pedestrian bounding boxes using the outermost edges of the labeled person instance segmentations, and we used the semantic segmentation of the road to define the SC region in the image.  For Cityscapes, the scene type is known to be ``city street''.  

We make Network Generalization Predictions for the JAAD and Cityscapes Datasets using $g(\textbf{c}^{S_K})$, estimated using the BDD Test Set. The prediction bias is consistently around $10\%$, with a minimum prediction error of $5\%$ for SC pedestrian recall in the JAAD Dataset. We underpredict pedestrian recall for the JAAD Dataset and we overpredict pedestrian recall for the Cityscapes Dataset. 

\section{Discussion}
\label{sec:discussion}

We make accurate Network Generalization Predictions for the BDD Operating Set, where the observed recall varies from $47\%$ to $87\%$. This demonstrates that a fixed test set can be used to predict a network's performance in diverse, novel operating domains. The observed recall for SC pedestrians is about $20\%$ higher than for NSC pedestrians. This makes intuitive sense, as SC pedestrians tend to be central in the image and closer to the vehicle. This is encouraging, because the performance of perception systems for autonomous vehicles will ultimately be determined by how well they detect SC pedestrians and obstacles. However, in the BDD Test Set there are many more examples of NSC pedestrians, $11,169$, than SC pedestrians, $484$. This leads to more untested contexts for the SC pedestrians, which in turn leads to the slight underprediction of SC recall. 

For unseen datasets, we find a Network Generalization Prediction bias of $10\%$; we believe these results are promising and that the results indicate the Context Subspace identified for one dataset, e.g., one camera setup and one physical setup, can be informative for unseen datasets. Investigating how network performance changes between datasets and identifying what physical changes lead to performance differences is a direction for future work. 

Network Generalization Prediction can be used to link network behavior in novel operating domains to required levels of performance. The Context Subspace can be leveraged for quasi-white box testing by testing the network across variations in context features that are known to impact network behavior. The Context Subspace also makes the network behavior interpretable by elucidating where failure is more likely.
In addition to making the Network Generalization Prediction tractable, we believe the Context Subspace can be used during network training  to extract features that are robust to changes in the Context Subspace. 
The Context Subspace can also be used for online error monitoring, e.g., an autonomous vehicle could notify the driver if it detects the surrounding scene is a context with subpar expected performance. 
We believe the Context Subspace is a tool that can make network performance more interpretable during training, testing, and deployment.


\section{Conclusions}
\label{sec:conclusion}

We propose the task Network Generalization Prediction and leverage a Context Subspace to render Network Generalization Prediction tractable with scarce test samples. We identify the Context Subspace  automatically and demonstrate accurate Network Generalization Prediction for Faster RCNN used for pedestrian detection in diverse operating domains. We show that the Context Subspace identified for the BDD Dataset is informative for completely unseen datasets. We believe that accurate Network Generalization Prediction, with an interpretable Context Subspace, is a step towards bridging the gap between the high performance of deep networks and the verification required for safety critical systems.


{\small
\bibliographystyle{ieee_fullname}
\bibliography{egbib}
}

\clearpage
\appendix

\section{Comparing $\Delta I$ and $I$}\label{app:derivation}
We propose $\Delta I(L, C^{s_1}, ..., C^{s_K})$ to approximate the Interaction Information between $K$ context features and the network loss $L$, $I(L, C^{s_1}, ..., C^{s_K})$. The computational complexity of computing $\Delta I(L, C^{s_1}, ..., C^{s_K})$ grows linearly with $K$, as compared to the computational complexity of computing $I(L, C^{s_1}, ..., C^{s_K})$ which grows combinatorially with $K$. We investigate the difference between $I(L, C^{s_1}, ..., C^{s_K})$ and $I(L, C^{s_1}, ..., C^{s_K})$. To simplify the notation, we denote $C^{s_1}$ as $C^1$ and $C^{s_2}$ as $C^2$. It is trivial to compute the Mutual Information between the context features and $L$ and select $C^1$ to be the feature most informative about the loss. We assume $C^1$ has been selected and we compare $I(L, C^1, C^2)$ and $\Delta I(L, C^1, C^2)$. 
\begin{equation}
I(L, C^1, C^2) = I(L, C^2) - I(L, C^2 | C^1)
\end{equation}
\begin{equation}
\Delta I(L, C^1, C^2) = I(L, C^2) - I(C^1, C^2)
\end{equation}
The difference between $I(L, C^1, C^2)$ and $\Delta I(L, C^1, C^2)$ is: 
\begin{equation}
I(L, C^1, C^2) - \Delta I(L, C^1, C^2) = I(C^1, C^2) - I(L, C^2 | C^1)
\end{equation}
As we would like the context features in $\textbf{C}^{S_K}$ to be roughly independent, let us assume that $C^1$ is not informative of $C^2$, i.e., $I(C^1, C^2) = 0$. 
\begin{equation}
I(L, C^1, C^2) - \Delta I(L, C^1, C^2) = - I(L, C^2 | C^1)
\end{equation}
The reader is reminded that the conditional mutual information is computed as:
\begin{multline}
I(L, C^{2}|C^{1})  = \sum_{\ell \in L} \sum_{c_1 \in C_1} \sum_{c_2 \in C_2} p(\ell, c_1, c_2) \\ 
\times  log \left( \frac{p(c_1)  p(\ell, c_1, c_2)}{p(\ell, c_1) p(c_1, c_2)} \right)
\end{multline}
For simplicity, let us consider the point wise conditional mutual information at $\ell$, $c_1$, and $c_2$:
\begin{equation}
  log \left( \frac{p(c_1)  p(\ell, c_1, c_2)}{p(\ell, c_1) p(c_1, c_2)} \right)
\end{equation}
Recall, it was assumed that $C^1$ and $C^2$ are independent, thus  $p(c_1, c_2) = p(c_1)p(c_2)$.
The joint probability $p(\ell, c_1, c_2)$ can also be factored as $\frac{1}{Z} \psi(\ell, c_1) \psi(\ell, c_2)$.
\begin{equation}
=  log \left( \frac{p(c_1)  \psi(\ell, c_1) \psi(\ell, c_2)}{Z p(\ell, c_1) p(c_1)p(c_2)} \right)
\end{equation}
\begin{equation}
=  log \left( \frac{  \psi(\ell, c_1) \psi(\ell, c_2)}{Z p(\ell, c_1) p(c_2)} \right)
\end{equation}
Note $ \psi(\ell, c_1) \propto p(\ell, c_1)$ and $ \psi(\ell, c_2) \propto p(\ell, c_2)$.
Thus, the difference between the proposed $\Delta I$ and the Interaction Information is proportional to 
\begin{equation}
 \propto log \left( p(\ell | c_2) \right)
\end{equation}
If we consider only combinations of $\ell$ and $c_2$ that exist in the test set, $p(\ell | c_2) > 0$. As the new context feature becomes more informative, $p(\ell | c_2) \rightarrow 1$ and the difference $log \left( p(\ell | c_2) \right) \rightarrow 0$. This demonstrates that, if the context features are informative about the loss, $\Delta I$ is a good approximation of the Interaction Information.

\end{document}